\documentclass{easychair}
\usepackage{cite}
\usepackage{url}
\usepackage{hyperref}
\usepackage{doi}
\usepackage{amsmath,amssymb,amsfonts}
\usepackage{graphicx}
\usepackage{lipsum}  
\usepackage{textcomp}
\usepackage{xcolor}
\usepackage{caption}
\usepackage{subcaption}
\usepackage{tikz}
\usepackage{float}
\usepackage{booktabs} 
\usepackage{multirow}
\usepackage{array}
\usetikzlibrary{arrows.meta, positioning}

\tikzset{
  box/.style={rectangle, draw, rounded corners, minimum width=0cm, minimum height=1cm, align=center, font=\small},
  arrow/.style={-Latex, thick},
}

\title{Data-Augmented Multimodal Feature Fusion for Multiclass Visual Recognition of Oral Cancer Lesions}

\author{
Joy Naoum\inst{1} \and 
Revana Salama\inst{1} \and 
Ali Hamdi\inst{1}
}

\institute{
MSA University, Giza, Egypt\\
\email{\{joy.ehab, revana.magdy, alihamdif\}@msa.edu.eg}
}

\titlerunning{Data-Augmented Multimodal Feature Fusion for Oral Cancer Recognition}

\begin{document}

\maketitle

\begin{abstract}
Oral cancer is frequently diagnosed at later stages due to its similarity to other lesions. Existing research on computer-aided diagnosis has made progress using deep learning; however, most approaches remain limited by small, imbalanced datasets and a dependence on single-modality features, which restricts model generalization in real-world clinical settings. To address these limitations, this study proposes a novel data-augmentation–driven multimodal feature-fusion framework integrated within a (Vision Recognition)VR-assisted oral cancer recognition system. Our method combines extensive data-centric augmentation with fused clinical and image-based representations to enhance model robustness and reduce diagnostic ambiguity. Using a stratified training pipeline and an EfficientNetV2-B1 backbone, the system improves feature diversity, mitigates imbalance, and strengthens the learned multimodal embeddings. Experimental evaluation demonstrates that the proposed framework achieves an overall accuracy of 82.57\% on 2 classes, 65.13\% on 3 classes, and 54.97\% on 4 classes, outperforming traditional single-stream CNN models. These results highlight the effectiveness of multimodal feature fusion combined with strategic augmentation for reliable early oral cancer lesion recognition and serve as a foundation for immersive VR-based clinical decision-support tools..
\end{abstract}

\section{Introduction}
\label{sect:introduction}

Approximately 377 thousand new diagnoses and 177 thousand deaths due to oral cancer occur each year, and the disease is a significant health challenge worldwide [1]. The stage at which a disease is diagnosed will always be a key determining factor in the disease prognosis. The five-year survival rate is nearly 80\% when oral cancer is diagnosed in the early stage, and drops below 30\% when diagnosed in a late stage [2]. Complicating the clinical diagnosis of oral cancer is the presence of overlapping visual traits among benign, precancerous, and malignant lesions, causing significant worldwide health challenges due to misdiagnosis. Most traditional computer-aided diagnostic (CAD) systems focus solely on the binary classification problem of determining whether a lesion is malignant and then simply distinguishing the malignant tissue from normal, non-cancerous ones. Although this is a good starting point for early screening, it is not nearly enough for a practical system in which a physician attempts to differentiate among a lesion's several subtypes. To assist clinicians in overcoming the dilemma of diagnostic uncertainty, the augmentation of systems that can perform multi-class classification is a necessity. In areas such as medical imaging, Deep learning, and in particular, convolutional neural networks have had a significant impact in the fields of dermatology, radiology, and pathology [3].Transfer learning with pretrained models like Inception-v3, DenseNet-121, and EfficientNet outperforms feature engineering in feature extraction, especially with small datasets. However, the application of deep learning in oral cancer detection is limited, predominantly focusing on binary classification. Multiclass approaches face challenges due to limited and imbalanced datasets, leading to poor model generalization and biased predictions. To overcome the issues in oral lesion classification, we offer an integrated strategy for multi-class classification, which incorporates stratified dataset splitting, augmented and oversampled datasets. This aims to improve model performance through better dataset construction on 2 types of oral lesions. The experiment's goal is to systematically reduce class imbalance, boost intraclass variation, and prevent overfitting to improve precision and recall for both rare and common lesions. Our model achieved 82.57\% accuracy, surpassing the state-of-the-art methods. Experimental results showcase that augmentation and oversampling improved the classification of the under-represented groups of lesions, confirming their value in the context of imbalanced clinical imaging. This study is a multiclass classification pipeline for oral lesions, consisting of 2 classes, and integrating research prototypes with clinical diagnostic assistance. This proposed pipeline improves model generalization and strength in the face of sparse, imbalanced clinical datasets. It confirms the data-centric approach in advancing computer-aided oral cancer diagnosis, showing improved results over the best methods available.

\section{Related Work}
\label{sect:related_work}

Deep learning techniques have recently advanced the ability of computers to automatically classify oral lesions from photographs. First, Devindi et al. [4] used multimodal data but were limited by a small dataset. Al-Ali et al. [1] achieved multiclass classification with 2,072 images, yet dataset size remained a constraint. Warin et al. [2] focused on lesion localization, limiting broader classification, while subsequent studies [3], [5] improved segmentation but still faced dataset diversity challenges .Devindi et al. [4] built on this model by incorporating a multimodal pipeline that used smartphone captured oral cavity photographs and patient attributes like age, sex, and lifestyle. Their model achieved an accuracy of 81\% after training with 2,271 annotated images using the advanced Image Encoders MobileNetV3-Large and ResNet-50. Despite these advances, limitations in dataset size, class balance, and robustness to varied imaging conditions persist, motivating the present study to leverage EfficientNetV2-B1 with augmented and stratified datasets for improved oral lesion classification. As much as these results illustrate the improvement of classification with the addition of contextual patient data, they also show the improvement of the clinically valuable tool in differentiating between benign and malignant oral lesions. This study also represented a departure from the majority of previous work on this problem, which focused on multimodal classification and used expensive imaging modalities. This study also continues in the same spirit, wherein the implementation of EfficientNetV2-B1, with stratified sampling, augmented target class balancing, and class balance training, is expected to provide improvements in diagnostic accuracy and robustness. \\
\\
 Additionally, Al-Ali et al. [1] built on this work with the creation of CLASEG, a multiclass classification system based on the EfficientNet-B3 architecture and trained with 2,072 clinical images of 14 distinct lesions. 74.49\% of classification accuracy was achieved, validating the clinical usefulness of consumer images to differentiate between lesions that are benign, premalignant, and malignant. Warin et al. [2] also contributed to this field by demonstrating that deep CNNs, in this case DenseNet-169, can attain a level of performance comparable to that of an expert in the detection of normal, potentially malignant, and oral squamous cell carcinoma tissues. Their study also investigated the object detection techniques Faster R-CNN and YOLOv5 in the context of localizing the lesions of interest. reinforcing the role of AI in supporting clinical decision-making. Subsequent work [3], [5] combined classification with lesion segmentation to achieve more precise boundary detection and improved differentiation between lesion subtypes. Collectively, the research exemplifies the shifting focus in the field of oral cancer screening toward more automated, adaptable, and comprehensive multiclass and multi-task deep learning approaches.\\
 \\
 Das et al. [6] was the first to address a completely separate area of research and deploy a deep learning technique for the multiclass classification of epithelial tissue cells in oral squamous cell carcinoma. Using transfer learning and CNNs with a training dataset achieving 80\% accuracy, they demonstrated the multiclass classification of highly heterogeneous histopathological images is indeed possible. Other studies describing the high-resolution automated image analysis and its implications for early cancer detection and the reduction of subjective errors in clinical diagnoses augment das et al and validate the importance of the analysis provided in the subsequent histopathological image classification of cancer lesions.However, the models used were relatively simple and may not fully capture the complex morphological variations in histopathological images, highlighting the need for more advanced architectures to improve feature extraction and classification performance.\\
 \\
 Albalawi et al. [7] also was granted the ability to use deep learning for a tissue-level classification task to perform the classification of histopathological images containing oral squamous cell carcinoma. They evaluated EfficientNet-B0 to B7 regarding a tissue dataset of 1,224 annotated digitized histopathological slides where EfficientNet-B7 was proved to outperform other classical CNNs including VGG-16, ResNet50 and achieved the peak accuracy of 96.3\% for the task. These results highlight the strength of compound scaling in capturing complex cellular structures and support the suitability of EfficientNet architectures for medical image analysis.Saldivia-Siracusa et al. [8] introduced a CNN-based framework for distinguishing oral potentially malignant disorders (OPMDs) from oral squamous cell carcinoma using 778 clinical photographs. Of the eight evaluated models, ConvNeXt and MobileNet (pre-trained on ImageNet and ISIC datasets) delivered the strongest performance, achieving 79.9\% accuracy, 83.7\% precision, 75.6\% recall, an F1-score of 79.4\%, and an AUROC of 86.3\%. Their results reinforce the effectiveness of CNNs and transfer learning techniques for classifying oral lesions captured under diverse imaging conditions in clinical settings.

\section{Methodology}
\label{sec:methodology}

\subsection{Dataset Description}

The study utilizes the Zenodo dataset \cite{ref4}, including 2,271 intraoral clinical photos of oral lesions captured in real-world conditions. These images were annotated with boundary and metadata features about the patients’ age, gender, smoking, alcohol consumption, and betel quid chewing habits. The images were originally divided into four categories: benign 748, oral cancer (129), healthy (729), and Oral Potentially Malignant Disorders (OPMD) 1394. The current study operated with the benign and OPMD categories, thus creating a binary classification problem with benign and OPMD as the two output classes. Such a restructuring of the problem statement led to a better distribution of the classes and as a result, better performance of the model.  To counter these class imbalances and avoid training the models with an unbalanced dataset, image augmentation and oversampling techniques were employed. For continuity and consistency compared to previous studies, the dataset was divided into training, validation, and testing splits following the methodology originally proposed by \cite{ref4}.

\begin{figure}[htbp]
\centering
\begin{subfigure}[b]{0.45\textwidth}
\centering
\includegraphics[width=\textwidth]{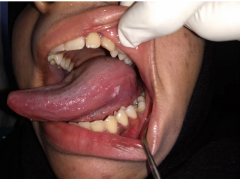}
\caption{Benign}
\end{subfigure}
\hfill
\begin{subfigure}[b]{0.45\textwidth}
\centering
\includegraphics[width=\textwidth]{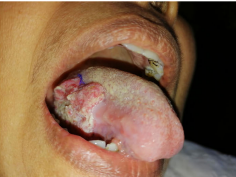}
\caption{OPMD}
\end{subfigure}
\caption{Sample images from the dataset showing different abnormal oral cavity regions (from Devindi et al. \cite{ref4}).}
\label{fig:dataset_samples}
\end{figure}

\subsection{Dataset PreProcessing}

Demographic attributes like age, gender, and habits like smoking, chewing betel quid, and alcohol drinking, this metadata was cleaned and normalized into numerical forms. For each patient ID, lesion photographs from the class-specific directories were gathered, resized to 224×224 pixels, and preprocessed with EfficientNetV2B1. In order to improve the model's resilience and mitigate the impact of the class imbalance, the training images were augmented. Further, pixel intensity values were also adjusted to be between 0 and 1, and standardized using the mean and standard deviation of the ImageNet, thereby allowing for more effective convergence and stable processes during the optimization. Each image pixel (x) was normalized to the range [0,1] using the following:\\
Standardization with ImageNet statistics was used. \\
Applied channel-wise mean ($\mu$) and standard deviation ($\sigma$) of the training set.

\[
X_{\text{norm}} = \frac{x - \mu}{\sigma}
\]

Additionally, metadata of the patients, including age, gender, smoking history, and chew time on betel quid and alcohol usage, was gathered from CSV file and subjected to cleansing and normalizing methods that included numerical encodings of the datasets. Then collected and arranged image lesions from the CSV file based on patient file IDs.

\subsection{Stratified Dataset Split}

The used dataset was stratified, and split into training (70\%), validation (15\%), and test (15\%) subsets. This was done to maintain the distribution of classes over the subsets.

\subsection{Augmentation Pipeline}

To simulate real-life variance, a multi-step augmentation approach using the Albumentations library was applied. Each image in the training set underwent various random transformations, with each image being augmented 5 times to enhance results:

\begin{itemize}
\item Horizontal and vertical flips, Random 90° rotations
\item Shift, scale, and rotate operations
\item Brightness and contrast adjustments, Hue and saturation shifts
\item Random resized cropping (scale: 0.7--1.0)
\item Gaussian blur and CLAHE (Contrast Limited Adaptive Histogram Equalization)
\item Coarse dropout (max 8 holes, 30$\times$30 pixels)
\end{itemize}

\subsection{Oversampling for Minority Classes}

In the random augmentation stage, classes that had fewer than 4200 were oversampled through random duplications of that class, to support a more fair distribution of representation across training iterations.

\subsection{Multimodal Learning}

Multimodal learning concerns models that learn simultaneously from different forms of data (also referred to as modalities). For this work, there are two selected modalities, which are;

\begin{itemize}
\item \textbf{Visual:} raw pixel (oral images).
\item \textbf{Tabular:} patient metadata (e.g., Age, Gender, Smoking, Betel chewing, Alcohol).
\end{itemize}

Having both allows the model to not only learn from the images, but also from the relevant clinical metadata. This boosts the classification performance when different forms of data are included compared to using images alone, as the structured attributes can provide contextual information that are not present in the images.
\begin{figure}[htbp]
\centering
\centering
\includegraphics[width=\textwidth]{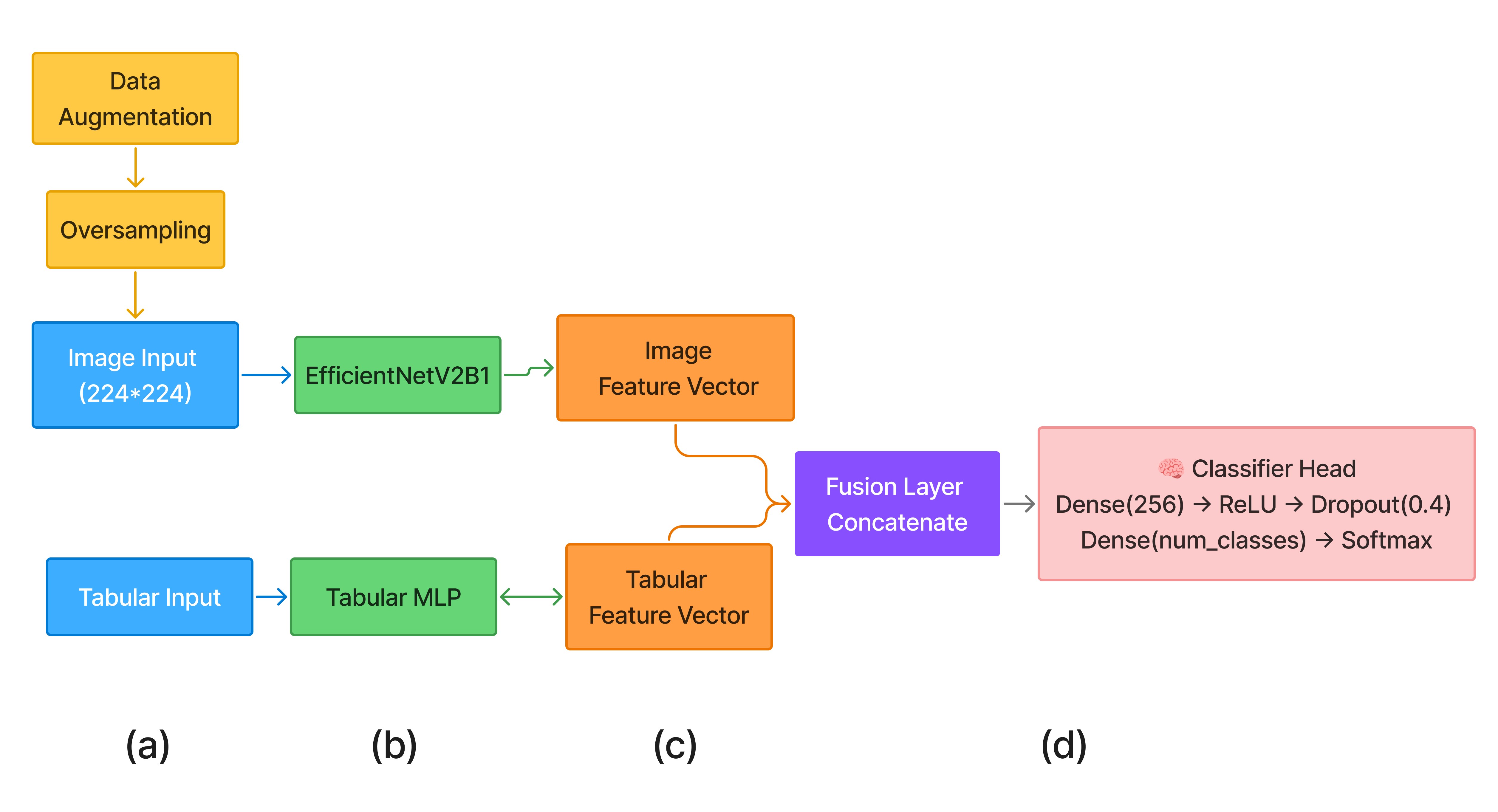}
\caption{Proposed Augmentation and Multimodal Fusion Workflow Architecture.}
\label{fig:augmentation_workflow}
\hfill

\end{figure}

\subsection{Fusion Architecture}
The overall model is composed of two branches:

\begin{itemize}
\item Image branch (deep CNN feature extractor)
\item Metadata branch (fully connected network for metadata)
\item Fusion layer (concatenation and joint processing)
\end{itemize}
To enhance model generalization and address the class imbalance, a custom augmentation pipeline was developed, specifically applied to the training set. The steps for augmentation include.

\subsection{Training and Optimization}

The feature extraction and fine-tuning phases marked two stages in the model’s training. The model was trained first at a learning rate of 1e-3 for 15 epochs, freezing the base layers and updating solely the top layers. For the fine-tuning phase, the bottom layers were partially unfrozen, and the learning rate was brought down to 1e-5 to refine representations. The layers were trained for another 10 epochs. The metrics were accuracy, precision, and recall, while the optimizer was Adam, and the loss was categorical cross-entropy. A batch size of 32 was maintained, and the passing of class weights combated class imbalance. Early Stopping, Model Checkpointing, and ReduceLROnPlateau were used among the optimization callbacks for stable convergence and overall best results in lesion classification for twelve lesion categories.

\subsection{Evaluation Metrics}

The model’s performance was evaluated on the test set using the following metrics: Overall Accuracy, Precision and Recall

\begin{table}[H]
\centering
\caption{Comparison of deep learning models on the test set}
\label{tab:model_comparison}
\makebox[\textwidth][c]{%
\begin{tabular}{lcccc}
\toprule
\textbf{Model} & \textbf{Img Size (px)} & \textbf{Batch Size} & \textbf{Epochs} & \textbf{Accuracy (\%)} \\
\midrule
DenseNet-121 & 224 & 64 & 200 & 0.74 \\
Inception-v3 & 224 & 64 & 200 & 0.76 \\
HRNet-W18-C & 224 & 64 & 200 & 0.79 \\
MixNet\_s & 224 & 64 & 200 & 0.77 \\
ResNet50 & 224 & 64 & 200 & 0.78 \\
MobileNetV3-Large & 224 & 64 & 200 & 0.81 \\
\textbf{Augmented MultiFusion CNN (EffNetV2-B1)} & 224 & 32 & 25 & \textbf{0.82} \\
\bottomrule
\end{tabular}%
}
\end{table}

\begin{table}[htbp]
\centering
\caption{Results on the test set of 4 classes only}
\label{tab:results_4_classes}
\begin{tabular}{lccc}
\toprule
\textbf{Classes} & \textbf{Precision} & \textbf{Recall} & \textbf{F1-score} \\
\midrule
OPMD & 0.6851 & 0.5905 & 0.6343 \\
Benign & 0.4615 & 0.5310 & 0.4938 \\
Healthy & 0.5660 & 0.5455 & 0.5556 \\
Oral Cancer & 0.1389 & 0.2500 & 0.1786 \\
\midrule
\multicolumn{3}{l}{\textbf{Overall Accuracy}} & \textbf{0.5497} \\
\bottomrule
\end{tabular}
\end{table}

\begin{table}[htbp]
\centering
\caption{Results on the test set of 3 classes only}
\label{tab:results_3_classes}
\begin{tabular}{lccc}
\toprule
\textbf{Classes} & \textbf{Precision} & \textbf{Recall} & \textbf{F1-score} \\
\midrule
OPMD & 0.747368 & 0.676190 & 0.710000 \\
Benign & 0.543307 & 0.610619 & 0.575000 \\
Healthy & 0.612069 & 0.645455 & 0.628319 \\
\midrule
\multicolumn{3}{l}{\textbf{Overall Accuracy}} & \textbf{0.6513} \\
\bottomrule
\end{tabular}
\end{table}

\begin{table}[htbp]
\centering
\caption{Results on the test set of 2 classes only}
\label{tab:results_2_classes}
\begin{tabular}{lccc}
\toprule
\textbf{Classes} & \textbf{Precision} & \textbf{Recall} & \textbf{F1-score} \\
\midrule
OPMD & 0.8326 & 0.8136 & 0.8230 \\
Benign & 0.9688 & 0.8378 & 0.8990 \\
\midrule
\multicolumn{3}{l}{\textbf{Overall Accuracy}} & \textbf{0.8257} \\
\bottomrule
\end{tabular}
\end{table}

\section{Results and Discussion}
\label{sec:results}

The proposed model, utilizing EfficientNetV2-B1 as a backbone, was trained and evaluated on a stratified dataset of two oral lesion categories using a two-step process: first training the classifier head with frozen base layers, followed by fine-tuning deeper layers at a slower learning rate. Overfitting was mitigated via early stopping, model checkpointing, and learning rate reduction. The model achieved 82.57\% overall accuracy, with 83.26\% precision and 81.36\% recall for the OPMD class, and 96.88\% precision and 83.78\% recall for the benign class, yielding high F1-scores (OPMD = 82.3, benign = 89.9), although it struggled with lesions sharing similar visual features. Confusion matrix analysis showed strong class separability despite some persistent misclassifications. Ablation studies confirmed that dataset enrichment through augmentation and oversampling improved performance significantly, boosting baseline accuracy from under 70\% to over 82\%. Compared to other CNN architectures, including DenseNet-121 (74\%), Inception-v3 (76\%), HRNet-W18-C (79\%), ResNet50, and MobileNetV3-Large, the proposed model outperformed them, benefiting from EfficientNetV2’s optimized depth, width, and resolution scaling to generate richer feature maps. Its low-parameter design and high inference speed make it suitable for clinical and low-resource environments. Limitations include a relatively small and imbalanced dataset, particularly under-representation of malignant cases, and untested variations in imaging conditions, which may affect generalization, sensitivity, and specificity. Future work should focus on expanding and diversifying datasets, incorporating multimodal data, and employing explainable AI approaches to enhance model transparency, clinical interpretability, and robustness.

\section{Conclusion}
\label{sec:conclusion}

In this work, we outlined a framework incorporating deep learning techniques that facilitates the automated detection and classification of oral lesions from clinical photographs. Utilizing a publicly available dataset with benign and malignant lesions across 2 classes, we showcased the potential of CDSS for implementation in clinical practice. To mitigate the effects of illumination variation, intra-class diversity, and class imbalance, systematic address of the oversight during the design of the preprocessing pipeline was incorporated, and included the normalization, augmentation, and oversampling of the dataset. From the class of models evaluated, our custom tuned Augmented MultiFusion CNN base model (EffNetV2-B1) out of all the recent deep learning architectures was able to surpass the previous record with a test accuracy of 82.57\%, and the highest accuracy on the test dataset compared to classical CNNs and all other EfficientNet derivatives. This effective use of transfer learning and model fine-tuning demonstrates the potential of deep learning for low resource medical imaging and underlines the importance of fine-tuning for transfer learning, within the context of smaller dataset sizes. The proposed method has the potential to act as a helpful diagnostic tool for doctors, enhancing early diagnosis of oral cancer and lowering diagnostic subjectivity. However, additional validation on bigger, more diverse datasets, as well as integration with clinical metadata, are required before application in real-world settings. Future study will concentrate on increasing the dataset, investigating multimodal approaches, and combining explainability methodologies to improve clinical trust and interpretability. Future work should expand datasets, integrate multimodal data, and use explainable AI for transparency. Lightweight models and knowledge distillation can enable deployment in low-resource settings.

\bibliographystyle{plain}
\nocite{*}\bibliography{easychair}

\end{document}